Original Paper

# Traditional Machine Learning Models and Bidirectional Encoder Representations From Transformer (BERT)–Based Automatic Classification of Tweets About Eating Disorders: Algorithm Development and Validation Study

José Alberto Benítez-Andrades[1*], BSc, MSc, PhD; José-Manuel Alija-Pérez[2], BSc, MSc, PhD; Maria-Esther Vidal[3*], BSc, MSc, PhD; Rafael Pastor-Vargas[4*], BSc, MSc, PhD; María Teresa García-Ordás[2*], BSc, MSc, PhD

[1]SALBIS Research Group, Department of Electric, Systems and Automatics Engineering, University of León, León, Spain
[2]SECOMUCI Research Group, Escuela de Ingenierías Industrial e Informática, Universidad de León, León, Spain
[3]Leibniz University of Hannover, Hannover, Germany
[4]Communications and Control Systems Department, Spanish National University for Distance Education, Madrid, Spain
[*]these authors contributed equally

**Corresponding Author:**
José Alberto Benítez-Andrades, BSc, MSc, PhD
SALBIS Research Group
Department of Electric, Systems and Automatics Engineering
University of León
Campus of Vegazana s/n
León, 24071
Spain
Phone: 34 987293628
Email: jbena@unileon.es

## Abstract

**Background:** Eating disorders affect an increasing number of people. Social networks provide information that can help.

**Objective:** We aimed to find machine learning models capable of efficiently categorizing tweets about eating disorders domain.

**Methods:** We collected tweets related to eating disorders, for 3 consecutive months. After preprocessing, a subset of 2000 tweets was labeled: (1) messages written by people suffering from eating disorders or not, (2) messages promoting suffering from eating disorders or not, (3) informative messages or not, and (4) scientific or nonscientific messages. Traditional machine learning and deep learning models were used to classify tweets. We evaluated accuracy, F1 score, and computational time for each model.

**Results:** A total of 1,058,957 tweets related to eating disorders were collected. were obtained in the 4 categorizations, with The bidirectional encoder representations from transformer–based models had the best score among the machine learning and deep learning techniques applied to the 4 categorization tasks (F1 scores 71.1%-86.4%).

**Conclusions:** Bidirectional encoder representations from transformer–based models have better performance, although their computational cost is significantly higher than those of traditional techniques, in classifying eating disorder–related tweets.

## Introduction

### Background

Physical appearance is an essential element for people in this society. Although many studies corroborate that moderate physical activity and proper nutrition help to maintain a healthy body [1] and mind [2], a large part of society continues to place more importance on physical appearance than on health. In recent years, trends have promoted a curvy physique [3,4] despite it being unhealthy, and most people associate having a slim body with being happy to have a slim body. These associations between physical appearance and happiness are the causes of illnesses such as eating disorders. These mental illnesses are complex and do not depend on a single factor [5,6]. Thus, messages relating being fat or skinny with aesthetics that are contained in some media—advertisements, magazines, and celebrity social media—can hurt people vulnerable to these types of illnesses.

The prevalence of eating disorders has been increasing [7]. In addition, since the start of the COVID-19 pandemic, there has been a more pronounced increase in eating disorders [7]. Therefore, any strategy that helps to combat this health problem may be of interest to society.

With the emergence of social media, studies [8-12] using social media data to propose solutions that can help combat this type of illness from different perspectives have also emerged. Artificial intelligence and machine learning techniques, mainly applied to text, have been used to find patterns that help in classifying text to explore eating disorder–related discourse shared on Twitter [8,9] and other studies [10-12] are making use of the data contained in social networks and offering solutions that can help in the field of public health. Among the social media currently available, the most widely used platform in scientific studies is Twitter [13].

Despite the increase in studies on eating disorders that have, for example, analyzed pro–eating disorder websites [14], performed sentiment analysis of pro–anorexia and antiproanorexia videos on YouTube [15], and that have used social media data and artificial intelligence techniques on pro–eating disorder and prorecovery text [16], none has identified (1) tweets that have been written by people who suffer or have suffered from eating disorders, (2) tweets that promote eating disorders, (3) informative or noninformative tweets related to eating disorders and, within the informative tweets, (4) which ones make use of scientific information and which ones do not.

### Objectives

Our main objectives were to achieve accurate text classification in performing these 4 tasks, to compare the efficiency of text classification models using traditional machine learning techniques and those using novel techniques, such as pretrained bidirectional encoder representations from transformer (BERT)–based models, to determine which approach has the best combination of performance and computational cost and would be useful for future research.

In our previous research [17], in which 6 test-beds were conducted, the main objective was only to apply 6 pretrained bidirectional encoder representations from transformer–based models to classify a category in a data set. This time, we used a broader approach, by presenting the main problem as a comparison of the performance (accuracy and computational cost) of traditional machine learning models vs bidirectional encoder representations from transformer–based models on 4 different data categorization tasks. This meant increasing 6 test-beds to 40 different test-beds.

### Literature Review

#### Social Media in Health Informatics

Social media, specifically social networks, have become very important sources of information within the field of health informatics. Health informatics includes the design and application of innovations based on information technologies to solve problems related to public health and health services [18]. In this branch of interdisciplinary science, it is possible to carry out complex research to manage information to improve efficiency and reduce costs in health care [19]. Health informatics includes information science, informatics, and health care.

Health-related research using social media is mainly focused on two areas. In real-time monitoring and the prediction of diseases (eg, influenza), it is possible to collect and use messages that have been geographically localized and that are on topics of interest. In this way, research tasks related to the user discussions are a simple task. Social media are also used to determine perspectives on different health problems and conditions. Thus, social media are useful, easy to use, and very important tools for observational studies.

Twitter is a very popular and widely used social network within the field of health and social health research. Some studies [9,14-16] make use of artificial intelligence techniques, such as social network data mining, to generate predictive models based on current knowledge. These techniques have been used, for example, in the context of the COVID-19 pandemic, to determine the public's perceptions [20] and to examine communication behavior between health organizations and users [21].

Studies [8,9,17,22-24] in the field of health informatics have used Twitter to study user behaviors and characteristics such as location, frequency, most used hashtags, or the structure of user networks. This information, being public and anonymous, is typically exempt from requiring the approval of an ethics committee [25]. Other studies have analyzed the impact of content shared by users [21] and how Twitter is used to receive and provide emotional support [26] or to determine the best way raise public awareness (World Rare Disease Day [27]).

Social media facilitate a great deal of research in the field of health informatics, for example, sentiment analysis, behavioral analysis, or information dissemination analysis, which make use of techniques related to machine learning or deep learning techniques for the classification and prediction of content that has been prepared using natural language processing.

### Classification Methods and Health Informatics

Supervised machine learning techniques are used to predict an outcome based on a given input by constructing an input–output pair. The main goal is to build a model that can then be used to make accurate predictions using new data.

Tasks in the field of supervised machine learning include regression—the prediction of a real number—and classification—the prediction of a class label [28]. Supervised classification tasks make use of a labeled training data set. This set allows the creation of classifiers or predictive models [28]. Text mining techniques are used to quantify text data (what is feature engineering) to represent the relationships between words as tokens.

Classification techniques make it possible to categorize large data sets efficiently to study text-based data. This approach has many advantages—more accurate predictions than those of humans and time savings [29-31]. Some commonly used classification techniques in health informatics are logistic regression, support vector machines, Naïve Bayes, random forest, gradient boosting trees, decision trees, and gradient boosted regression trees.

Naïve Bayes classifiers have been used to predict Zika and dengue diseases using data obtained from Twitter [32] and to test the classification of 4 conditions—influenza, depression, pregnancy, and eating disorders—and 2 locations—Portugal and Spain [33].

Other studies [34-36] have shown that good results can be obtained using support vector machine algorithms, such as, with a neural network to predict COVID-19 in chest x-ray images, a prediction model [35], and for sentiment analysis tasks on a Twitter data set related to the COVID-19 pandemic in Canada [36]. A gradient boosted regression tree classifier was used to identify tweets related to e-cigarettes [22], with accurate classification of 5 different user types, by manually labeling a sample of tweets and using feature engineering techniques based on the term frequency–inverse document frequency matrix.

It is also possible to combine different classification algorithms and compare their performance to use the best performing classifier for a given task [37], for example, gradient boosting tree, decision tree, logistic regression, and support vector machine models were used to predict patient needs at the level of informational support [23].

### Social Media Research Related to Eating Disorders

There are a number of studies that make use of data related to eating disorders [8,14-16,24,38]. In one study [24], 123,977 tweets were collected and a subsample of 2219 was labeled; the efficiency of a convolutional neural network, with long short-term memory, in classifying tweets about eating disorders was demonstrated. Another study [8] statistically analyzed the effect of eating disorder awareness campaigns by obtaining information on tweets that mentioned 2 hashtags. A review [39] showed the importance of machine learning in advancing the prediction, prevention, and treatment of mental illness and eating disorders. Other studies [14-16,38] have demonstrated the importance of the use of data obtained from social networks in the field of eating disorders, by performing analysis from a social rather than computational perspective, which is known as social network analysis.

A previous social media study predicted depression from texts [40]; therefore, detecting texts written by people suffering from eating disorders can also be helpful. Studies on the detection of pro-ana and pro-recovery communities [41-43]—people in favor of and who promote anorexia and recovery from eating disorders, respectively—and reviews [44,45], have suggested this type of study may be useful. Furthermore, to the best of our knowledge, no studies having the same objectives as ours have been conducted.

## Methods

### Data Collection

#### Tweets

A tool (T-Hoarder [46]) was used to collect tweets (Figure 1). Tweets were obtained at the moment they were sent because the tool uses the Twitter streaming API, thus tweets that were subsequently removed from the platform for not complying with regulations were still obtained.

T-Hoarder allowed us to obtain additional information about tweets for further analysis, such as, ID, text, and author (among other fields). Tweets were identified by keywords [17]. In set 1 "anorexia," "anorexic," "dietary disorders," "inappetence," "feeding disorder," "food problem," "binge eating," and "anorectic" we used. In set 2, "eating disorders," "bulimia," "food issues," "loss of appetite," "food issue," "food hater," "eat healthier," "disturbed eating habits," "abnormal eating habits," and "abnormal eating habit" were used. In set 3, "binge-vomit syndrome," "bingeing," "bulimarexia," "anorexic skinny," and "eating healthy" were used.

By using a different Twitter accounts for each set, more tweets could be obtained without exceeding the Twitter platform's usage limit. English terms were used because more tweets are generated in English [46].

**Figure 1.** Study workflow: (A) data collection and preprocessing, (B) classification model training, and (C) evaluation. BERT: bidirectional encoder representations from transformer, ML: machine learning.

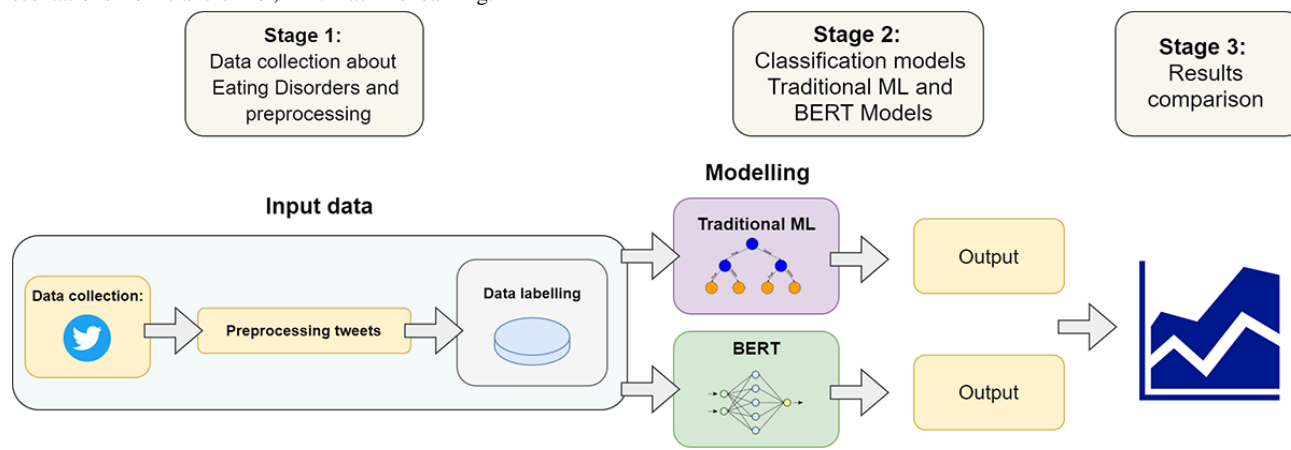

## Preprocessing

Preprocessing was conducted in Python (version 3.6). Data were loaded from documents obtained through T-Hoarder, which generates a file up to 100 MB; therefore, 4 files were obtained for data set 1, 4 files were obtained for data set 2, and 2 files were obtained for data set 3. Some data, such as location, name, and biography, contained line breaks or tabs. To avoid conflicts with delimiters, tabs and line breaks were removed using a function. After preprocessing the data frames, they were concatenated into a single data frame. In order to be able to work in a more agile way with the data frame, the memory usage of the data frame was calculated and optimized by converting numeric columns into numbers, converting dates to datetime format, and converting the remaining objects into categories. These steps helped reduce the data frame from 2.7 GB to 1.1 GB. We removed all tweets that were retweets, duplicates (because we unified data sets that might contain common tweets), and non–English tweets.

To select the subset of 2000 tweets, manual filtering was performed to eliminate tweets that were not related to eating disorder issues. Some of our keywords were too generic and meant that the tweets collected were not about eating disorders. For example, some of these words that triggered the collection of tweets unrelated to eating disorders were "food problem," "inappetence," "food issue," and "bingeing"; however, in order to generate predictive models with greater accuracy and less bias, we kept a small sample of tweets (n=286) that did not belong to any of the categories, but that did contain some of the keywords of interest.

## Labeling

Tweets in 4 different categories in the subset were manually labeled ([Table 1](#)). Labeling was carried out by 2 people, labeling 1000 tweets each. The labels were then reviewed by 4 mental illness experts. This procedure took place over the course of 1 full month, with each person taking approximately a total of 70 hours in carrying out this work. In category 1, tweets written by people suffering from eating disorders were represented with a value of 1, and the rest were represented with a value of 0. To assess this, each user profile was accessed and user description and tweets published by the user were examined to determine if the user had publicly mentioned having an eating disorder. In category 2, tweets that promoted having an eating disorder were labeled with a value of 1, and all other tweets were labeled with a value of 0. There are communities of people who suffer from eating disorders who try to encourage other people to also suffer from it by promoting it as if it were something positive or fashionable. There are many studies [9] that talk about pro–eating disorders communities using the terms *pro-ana* or *pro-anorexia*. In category 3, informative tweets were represented with a value of 1, and noninformative tweets were represented with a value of 0. Informative tweets are those that show information with the aim of informing readers, while the rest were those in which the author expressed an opinion. In category 4, scientific tweets were labeled with a value of 1, and the rest were labeled with a value of 0. A tweet of an informative nature that had been written by a person belonging to the field of research, for example, a doctor of philosophy in different subjects, was labeled as a scientific tweet. Scientific tweets were also those that shared links to papers published in scientific journals. If a tweet did not belong to any of the 4 proposed categories, it was not eliminated from the data set, since having tweets with value of 0 was also necessary.

Table 1. Categories of labeled tweets and examples.

| Category topics | Tweet |
| --- | --- |
| **Category 1** | |
| Written by someone who suffers from eating disorder | i was stressed and ate a whole bowl of pasta, where's my badge for being the worst anorexic #edtwt |
| Written by someone who does not have an eating disorder | Is your #teenager not eating or eating a lot less than normal? She might be suffering from #anorexia. We can help; please come see us https://t.co/GfStM1IVGz #weightloss #losingweight https://t.co/z5NK0tjNIt |
| **Category 2** | |
| Promotes eating disorders | Currently feeling like the best anorexic #eating disordertwt https://t.co/1BZPMs8bGU #mentalhealth #diet #anorexia |
| Not promotes eating disorders | Higher-calorie diets could lead to a speedier recovery in patients with anorexia nervosa, study shows https://t.co/mipX3nrhHN |
| **Category 3** | |
| Informative | #AnorexiaNervosa – A Father and Daughter Perspective -Highlights from RCPsychIC 2019 # EatingDisorders #mentalhealth https://t.co/iq3GH5ce6C |
| Noninformative | Binge eating makes me sad :( #eatingdisorder #bingeeating https://t.co/0jjf7YrVyc |
| **Category 4** | |
| Scientific | The problem extends to Food and Drug Administration and National Institutes of Health data sets used in a recent study appearing in Reproductive Toxicology. #ai #technology #BigData #ML https://t.co/DFvh6gNA38 |
| Nonscientific | Do not waste time thinking about what you could have done differently. Keep your eyes on the road ahead and do it differently now. #anorexia #eatingdis- order #recovery #nevergiveup #alwayskeepfighting https://t.co/YalYzclBDM |

### Final Sample

Before training and validating the models, tweets in the labeled set with more than 80% similarity were eliminated. It was decided to apply this criterion for tweets containing the same text but using different hashtags. Remaining tweets were processed by removing the stop words (words that have no meaning on their own and that modify or accompany other words, for example, articles, pronouns, adverbs, prepositions, or some verbs) and punctuation or symbols, that hindered the application of machine learning techniques.

### Classification Methods

#### General

We used random forest, recurrent neural networks, bidirectional long short-term memory networks (ClassificationModel; simpletransformers [47], version 0.62.2), and pretrained bidirectional encoder representations from transformer–based models (RoBERTa [48], BERT [49], CamemBERT [50], DistilBERT [51], FlauBERT [52], ALBERT [53], and RobBERT [54]). Bidirectional long short-term memory and bidirectional encoder representations from transformer–based methods were chosen because they seemed to be the most promising models for natural language processing [55-57]. In addition, random forest was used for comparison because it is a traditional machine learning technique.

Two models—CamemBERT [50] and FlauBERT [52]—were pretrained using French text, and RobBERT [54] was pretrained using Dutch text. We used these models to obtain performance data for with text not written in their initial language. Data were divided into 70% training and 30% testing sets (train_test_split function in scikit-learn). The evaluation metrics were accuracy and F1 score.

For the random forest model, 5-fold cross-validation was used. For the neural networks, 5 different iterations were performed, and the mean F1 score and accuracy were obtained.

#### Random Forest

Random forest models [58] are constructed from a set of decision trees, which are usually trained with a method called bagging, to take advantage of the independence between the simple algorithms, since error can be greatly reduced by averaging the outputs of the simple models. Several decision trees are built and fused in order to obtain a more stable and accurate prediction. Random forest models can be used for both regression and classification problems.

One of the advantages offered by this type of model is the additional randomness when more trees are included. The algorithm searches for the best feature as a node is split from a random set of features. This makes it possible to obtain models with better performance. When a node is split, only a random subset of features is considered. Random thresholds can also be used for each feature, instead of searching for the best possible threshold, which adds additional randomness.

#### Recurrent Neural Network

In this type of neural network, a temporal sequence that contains a directed graph made up of connections between different nodes is defined. These networks have the capacity to show a dynamic temporal behavior. These types of networks, which are derived from feedforward neural networks, have the ability to use memory (their internal state) to process input sequences of varying lengths. This feature makes recurrent neural networks useful for tasks such as unsegmented and connected handwriting recognition or speech recognition [55,59,60].

There are 2 classes of recurrent neural networks—finite-pulse and infinite-pulse. The former are made up of a directed acyclic graph that can be unrolled and replaced by a strictly feedforward neural network, whereas the latter are made of a directed cyclic graph, which does not allow the graph from being unrolled.

### Bidirectional Long Short-Term Memory

Bidirectional long short-term memory networks [61] are constructed from 2 long short-term memory modules that, at each time step, take past and future states into account to produce the output.

### Bidirectional Encoder Representations From Transformer–Based Models

The bidirectional encoder representations from transformer framework is not a model in itself. According to Devlin et al [49], it is a "language understanding" model.

In the bidirectional encoder representations from transformer–based method, a neural network is trained to learn a language, similar to transfer learning in computer vision neural networks, and follows the linguistic representation in a bidirectional way, looking at the words both after and before each words. It is the combination of these approaches that has made it a successful natural language processing method [62].

### Configuration

We used Jupyter notebook and TensorFlow and Pytorch libraries. It was necessary to use both libraries because, currently, bidirectional encoder representations from transformer–based networks can only be generated through Pytorch, while TensorFlow is one of the most widely used libraries to generate random forest, recurrent neural network, and bidirectional long short-term memory models.

### Hyperparameters

We used a grid search (GridSearchCV) to select the random forest parameters (Table 2).

To train recurrent neural networks (sklearn; keras) to perform the binary categorization tasks, the sigmoid activation function used (Figure 2). We trained and validated the bidirectional long short-term memory models (sklearn and TensorFlow libraries) using the best-performing configuration (Figure 3), after carrying out different tests.

For the 7 pretrained bidirectional encoder representations from transformer–based models, the hyperparameters were *reprocess_input_data*=True; fp16=False; *evaluate_during_training*=False; *evaluate_during_training_verbose*=False; *learning_rate*=2e-5; *train_batch_size*=32; *eval_batch_size*=32; *num_train_epochs*=15; *overwrite_output_dir*=True; and *evaluation_strategy*='epochs'.

All experiments and data are published in a repository accessible to anyone [63].

Table 2. Random forest hyperparameters.

| Category | criterion | max_depth | max_features | n_estimators |
|---|---|---|---|---|
| Category 1 | gini | 7 | log2 | 200 |
| Category 2 | gini | 8 | auto | 1000 |
| Category 3 | gini | 8 | sqrt | 800 |
| Category 4 | gini | 8 | auto | 1000 |

**Figure 2.** Architecture of the recurrent neural network network. LSTM: long short-term memory.

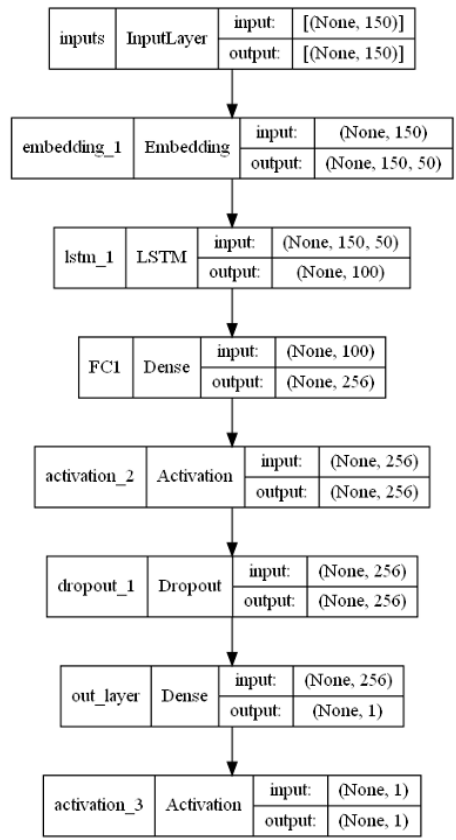

**Figure 3.** Architecture of the bidirectional long short-term memory (LSTM) network.

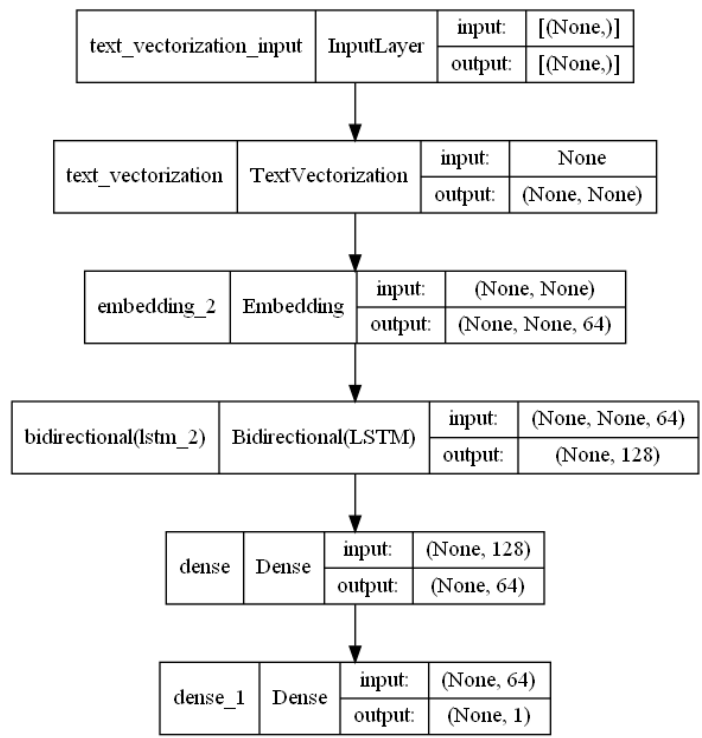

## Results

### Preprocessing

A total of 1,085,957 tweets, written and posted on Twitter between October 20, 2020 and December 26, 2020, were collected. After preprocessing, a total of 494,025 valid tweets were obtained. These tweets are shared and publicly available on the Kaggle platform [64]. From the subset of 2000 tweets that was manually labeled, 1877 remained after the similarity criterion was applied. Table 3 shows the 10 most repeated terms in the full set of tweets and in the subset that was labeled.

**Table 3.** Table of terms and frequencies of the 10 most repeated terms in the initial data set and in the labeled subset of data.

| Term | Frequency, n |
| --- | --- |
| **Complete set (n=494,025)** | |
| hey mp | 230,013 |
| healthy | 210,430 |
| pltpinkmonday | 209,330 |
| eat | 183,436 |
| covid19 | 156,541 |
| edtwt | 123,175 |
| anorexia | 112,864 |
| disorders | 102,063 |
| endsars | 99,844 |
| bachelorette | 48,370 |
| problem | 45,959 |
| **Subset (n=2000)** | |
| eat | 1132 |
| disorder | 830 |
| food | 410 |
| recovery | 382 |
| edtwt | 301 |
| binge | 282 |
| people | 245 |
| anorexic | 244 |
| research | 226 |
| study | 202 |
| problem | 199 |

### Category

In category 1, 50.2% (942/1877) of tweets were written by a person with an eating disorder, and 49.8% (935/1877) of tweets were written by a person without an eating disorder. In category 2, 23.8% (447/1877) of tweets encourage people to suffer from an eating disorder, and 76.2% (1400/1877) of tweets do not encourage people to suffer from an eating disorder.

In category 3, 37% (694/1877) of the tweets were informative, 63% (1183/1877) of tweets were opinionated. In category 4, 23.3% (437/1877) of the tweets were scientific, 76.7% (1440/1877) of tweets were of a nonscientific nature.

### Performance

Performance (Table 4) and implementation time (Table 5), which corresponds to the time invested in generating and validating the different models, for 4 different categorization tasks. The pretrained RoBERTa model was the most accurate for detecting tweets that had been written by people suffering from some type of eating disorder (accuracy 83.1%). Despite this, the more traditional recurrent neural network yielded an accuracy that was not much lower (accuracy 82.6%). The most accurate model for the detection of tweets that did or did not promote an eating disorder was the RoBERTa model (accuracy 88.5%); however, applying bidirectional long short-term memory improved performance (accuracy 86.7%). The most accurate model for the detection of informative or opinion-based tweets was RoBERTa model (accuracy 84.4%). Accuracy for all bidirectional encoder representations from transformer–based models, except ALBERT and FlauBERT, exceeded 80%; however, applying bidirectional long short-term memory resulted in an accuracy of 78.7%. The model with the highest accuracy for the detection of scientific or nonscientific tweets was the RoBERTa model (accuracy 94.2%). All bidirectional encoder representations from transformer–based models equaled or exceeded 92%; however, applying bidirectional long short-term memory yielded an accuracy of 85.8%.

**Table 4.** Classification performance.

| Model | Having eating disorders or not | | Encouraging eating disorders or not | | Informative or not | | Scientific or not | |
|---|---|---|---|---|---|---|---|---|
| | F1 score, % | Accuracy, % | F1 score, % | Accuracy, % | F1 score, % | Accuracy, % | F1 score, % | Accuracy, % |
| Random forest | 79.8 | 79.2 | 47 | 76.7 | 49.2 | 73.7 | 27.3 | 80.4 |
| Recurrent neural network | 83.2 | 82.6 | 61 | 82.1 | 67.3 | 70.7 | 67.3 | 70.7 |
| Bidirectional long short-term memory | 78.5 | 79.3 | 67.1 | 86.7 | 67.1 | 78.7 | 76.8 | 85.8 |
| Bidirectional encoder representations from transformer–based[a] | 83.3 | 83 | 71.9 | 87.2 | 77.6 | 84.3 | 86 | 94.1 |
| RoBERTa[a] | 83.8 | 83.1 | 74.3 | 88.5 | 77.6 | 84.4 | 86.4 | 94.2 |
| DistilBERT[a] | 84 | 83.1 | 72.3 | 87.3 | 75 | 82.8 | 84.2 | 93.3 |
| CamemBERT[a] | 79.1 | 78.7 | 73.6 | 87.8 | 74.7 | 81.7 | 82.5 | 92.3 |
| ALBERT[a] | 81.2 | 80.4 | 74.3 | 88.2 | 73.8 | 81.5 | 83.3 | 93 |
| FlauBERT[a] | 82.6 | 81.7 | 72.9 | 87.5 | 72.2 | 80 | 83.4 | 92.7 |
| RobBERT[a] | 78.8 | 78.4 | 71.1 | 86.2 | 73.8 | 81.6 | 83 | 92.6 |

[a]A pretrained model was used: bert-based-multilingual-cased for BERT, roberta-base for RoBERTa, distilbert-base-cased for DistilBERT, camembert-base for CamemBERT, albert-base-v1 for ALBERT, flaubert-base-cased for FlauBERT, and robbert-v2-dutch-base for RobBERT.

For bidirectional encoder representations from transformer–based models, despite obtaining better performance metrics in terms of accuracy, the training and validation times of the models are much higher than those of random forest, recurrent neural network, and bidirectional long short-term memory models. For example, bidirectional encoder representations from transformer–based models take approximately 15 times longer than random forest models (Table 5).

The improvements between the accuracy of the best bidirectional encoder representations from transformer–based model (Categorization 1: DistilBERT 83.1%; Categorization 2: RoBERTa 88.5%; Categorization 3: RoBERTa 84.4%; Categorization 4: RoBERTa 94.2%) and that of the best model between random forest, recurrent neural network, or bidirectional long short-term memory models (Categorization 1: recurrent neural network 82.6%; Categorization 2: bidirectional long short-term memory 86.7%; Categorization 3: bidirectional long short-term memory 78.7%; Categorization 4: bidirectional long short-term memory 85.8%) were 0.61%, 2.08%, 7.24%, and 9.79%, respectively.

**Table 5.** Implementation time.

| Model | Time (seconds) | | | |
|---|---|---|---|---|
| | Having eating disorders or not | Encouraging eating disorders or not | Informative or not | Scientific or not |
| Random forest | 1.74 | 12.8 | 10.4 | 12.9 |
| Recurrent neural network | 152.1 | 163.1 | 151.5 | 153.7 |
| Bidirectional long short-term memory | 163.2 | 175.3 | 164.8 | 167.9 |
| Bidirectional encoder representations from transformer–based | 1257.4 | 1232.1 | 1292.7 | 1311.4 |
| RoBERTa | 1116.2 | 1158.8 | 1142.5 | 1192.8 |
| DistilBERT | 1343.3 | 1327.8 | 1332.0 | 1362.3 |
| CamemBERT | 1472.3 | 1457.5 | 1462.0 | 1493.4 |
| ALBERT | 1372.7 | 1352.3 | 1331.3 | 1392.5 |
| FlauBERT | 1203.9 | 1207.1 | 1202.1 | 1235.1 |
| RobBERT | 1234.4 | 1215.4 | 1319.7 | 1123.5 |

## Discussion

### Principal Results

Practitioners and researchers can benefit from the use of social media data in the field of eating disorder. Although the model with the best accuracy was always one of the pretrained bidirectional encoder representations from transformer–based models, the computational costs compared with those of simpler models may be excessive. The difference between the accuracy of the best bidirectional encoder representations from transformer–based model and the best of the 3 simpler models (random forest, recurrent neural network, and bidirectional long short-term memory) did not exceed 9.79%.

Given the high computational cost, use of bidirectional encoder representations from transformer–based models in this instance may not be essential. The accuracy for the 4 different categorization tasks is relatively high even in the simplest models.

Despite the fact that we used only 1877 tweets (which is similar to the amounts used in previous studies: 2219 [24] and 2095 [65]), the models classified the tweets with a high level of accuracy.

For the classification of tweets into informative or noninformative (categorization 3), our models obtained a higher accuracy (80%-84.4%) than those in previous studies (77.7% [44] and 81% [45]). Comparisons cannot be made for the other 3 categorization tasks because of a lack pf applicable eating disorder–related studies.

### Limitations

This research has several limitations. (1) It was limited to a social media platform, (2) some categorization tasks were not balanced, which may lead to bias in the generated models, (3) the training set was sufficient but could be larger for better results in a real environment, and (4) when labeling tweets, it is possible that there was a bias in determining whether a tweet was written by someone with an eating disorder due to lack of information about the user.

### Conclusions

Machine learning and deep learning models were used to classify eating disorder–related tweets into binary categories in 4 categorization tasks, with accuracies greater than 80%. The best performing models were RoBERTa and DistilBERT, both bidirectional encoder representations from transformer–based classification methods.

The computational cost was much higher for the bidirectional encoder representations from transformer–based models compared to those of the simpler models (random forest, recurrent neural network or traditional bidirectional long short-term memory), time invested in training and validation was greater by a factor of 10.

Future work will include (1) increasing the training and validation data set, (2) applying natural language processing techniques that make use of ontologies with which it is possible to include automation and logical reasoning, (3) integrating predictive models in a real-world development project, such as a Twitter bot, and (4) validating the model using texts written by patients with eating disorders and who are in treatment.

### Conflicts of Interest

None declared.

XSL•FO
RenderX